# Exploratory Simulation of an Intelligent Iris Verifier Distributed System

N. Popescu-Bodorin*, *Member IEEE* and V.E. Balas**, *Senior Member IEEE*
* Artificial Intelligence & Computational Logic Laboratory, *Spiru Haret* University, Bucharest, România
** Faculty of Engineering, Aurel Vlaicu University, Arad, România
bodorin@ieee.org, balas@drbalas.ro

*Abstract* — This paper discusses some topics related to the latest trends in the field of evolutionary approaches to iris recognition. It presents the results of an exploratory experimental simulation whose goal was to analyze the possibility of establishing an Interchange Protocol for Digital Identities evolved in different geographic locations interconnected through and into an Intelligent Iris Verifier Distributed System (IIVDS) based on multi-enrollment. Finding a logically consistent model for the Interchange Protocol is the key factor in designing the future large-scale iris biometric networks. Therefore, the logical model of such a protocol is also investigated here. All tests are made on Bath Iris Database and prove that outstanding power of discrimination between the intra- and the inter-class comparisons can be achieved by an IIVDS, even when practicing 52.759.182 inter-class and 10.991.943 intra-class comparisons. Still, the test results confirm that inconsistent enrollment can change the logic of recognition from a fuzzified 2-valent consistent logic of biometric certitudes to a fuzzified 3-valent inconsistent possibilistic logic of biometric beliefs justified through experimentally determined probabilities, or to a fuzzified 8-valent logic which is *almost consistent* as a biometric theory - this quality being counterbalanced by an absolutely reasonable loss in the user comfort level.

## I. Introduction

The evolutionary approach to iris recognition [1] is a very recent topic, indeed. The study of Consistent Biometry [1] and the study concerning the logical consistency of iris recognition ([1], [2]) are also new research directions. All of these three topics came from a different perspective of iris recognition, which is considered a problem of computational logic and artificial intelligence, a hypostasis of the more general problem of logical and intelligent understanding of data.

Daugman introduced the classical statistical perspective of iris recognition [4], [5] and many others [6]-[13] followed his view. The difference between the classical statistical decision landscape of iris recognition and the evolutionary model of iris recognition is explained in [1].

The present paper extends and uses the results previously presented in [1], [2] and [3] by analyzing the possibilities of establishing an Interchange Protocol for Digital Identities evolved in different geographic locations interconnected through and into an Intelligent Iris Verifier Distributed System (IIVDS) based on multi-enrollment. The goal of such a study is finding a logically consistent model for the Interchange Protocol - the key factor in designing the future large-scale iris biometric networks.

### A. Terminology

An Intelligent Iris Verifier (IIV, [1]) is a non-stationary, complex, and logically self-aware [1], [2] biometric system which knows Cognitive Binary Logic [2], Consistent Biometry [1] and custom arithmetic languages, all of these enabling it to preserve its logical consistency [1], [2], i.e. to overcome the pressure of the new enrollments through logical and intelligent evolution [1].

The Intelligent Iris Verifier Distributed System (IIVDS, N. Popescu-Bodorin) consists in multiple instances of IIV systems [1] interconnected into a large star-network topology that specifies the central unit (CU) and the terminal stations.

A minimal theory 'T' of iris recognition consists in a given vocabulary 'V' of binary iris codes, the digital identities, and a given knowledge 'K' about them (a grammar) describing well-formed (legal and meaningful) computation with elements of vocabulary: how to compute digital identities from a given number of binary iris codes and how to test the similarity between the digital identities and the binary iris codes:

$$T = (V, K).$$

Knowing that IIVDS is an evolutionary system [1], we see that even a minimal theory of iris recognition reflects a dynamic understanding of iris recognition based on the available experience at certain moment 't' and continuously evolves under the pressure of the new enrollments (the stress factor, [1]):

$$T_t = (V_t, K_t).$$

A multi-enrollment scenario in iris recognition is that in which a given number of hypostases of the same iris taken from the same person enroll in the system under the ID number of that person.

A positive/negative identity claim is something like "*I am/I'm not* the user X".

The details about digital identities, individual evolution, systemic evolution, geometrical meaning of evolving digital identities and more can be found in [1].

Also, the logical landscape (the formal theory of binary logic, Liar Paradox) in which the present paper is integrated is given in [1], and [2].

## II. Problem Formulation

Here in this paper we aim to analyze the following scenario:

- In an IIVDS based on multi-enrollment, the Central Unit (CU) evolves (discovers) a logically consistent theory of iris recognition, as described in [1] (see Fig. 3 from [1]; it is shown there that the CU proves a logically consistent understanding of iris recognition, which in this case is described by a fuzzified 2-valent consistent logic of biometric certitudes [1]), by practicing consistent enrollment for five binary iris codes per identity. This theory will be denoted as 'T' and describes how to extract meaningful information from a number of iris codes, how to assemble this information into a digital identity, and how to compare digital identities to binary iris codes. As we already said, $T = (V, K)$.

- The central unit of the IIVDS spreads its knowledge K through the entire IIVDS by distributing K to all terminals. For obvious reasons, this action will be further referred to as *centrifugal knowledge synchronization* or *centrifugal knowledge dissemination* (CKD) and describes the center-to-terminals knowledge flow.

- The terminals within the IIVDS receive and accept the grammar K that becomes the new grammar in the local theory of iris recognition on all terminals. The grammar K replaces without negotiation all grammars within the local theories evolved individually on each terminal, i.e. CKD is a mandatory knowledge update for all terminals.

- At this stage and until the end of the test, all stations within the IIVDS share the same knowledge (the same 'global' grammar) but CU allows all terminals to work on different data (local vocabularies). To simulate this situation, the CU distributes the database of iris codes (16x256 binary matrices generated with Haar-Hilbert encoder HH1 introduced in [3]) to all terminals but allows each terminal to practice random (instead consistent) enrollment with five binary iris codes per identity.

- All terminal stations within IIVDS will simulate individual evolutions [1] for all 50 digital identities available in the database. The quality of the learning is then tested on each terminal using those binary iris codes unseen by the local IIV during the learning (the local test dataset). Each terminal forwards the results obtained by comparing all enrolled identities to all candidate binary iris codes within the test dataset to the central unit.

- The central unit collects all statistics delivered by all terminals and interprets them.

Reference [1] explained the reasons for which the consistent enrollment leads to a fuzzified but consistent logical understanding of iris recognition (Fig. 3 from [1]). The main question here is that:

*Will be the Central Unit still capable to prove a logically consistent understanding of iris recognition after it summarizes the experience of the entire network?*

Naturally, a deterministic machine can only do what is designed to do, and therefore, the answer depends on how the CU is endowed for achieving the goal of interpreting experimental results. Two possibilities of endowing CU with a logical theory of interpretation will be examined here:

- In the first case, CU is assumed to know a 3-valent fuzzy logic of iris recognition in which the truth values are: fuzzy zero (two irides are different if their comparison returns this value), fuzzy 1 (two irides are similar if their comparison returns this value), and fuzzy undecidable – meaning that two irides have equal chances to be different and similar simultaneously, if their comparison returns this value (the case of the point of Equal Error Rate, [4]).

- In the second case, CU is assumed to know an 8-valent fuzzy logic of iris recognition generated by three seeds: *fuzzy 0* - the interval of similarity scores on which a Negative claim is Accepted (NA) and a Positive claim is Rejected (PR), *fuzzy 1* - the interval of similarity scores on which a Positive claim is Accepted (PA) and a Negative claim is Rejected (NR), *fuzzy uncertain* - the interval of similarity scores on which a Positive claim is Rejected (PR) and a Negative claim is Rejected (NR):

$$F0 \equiv PR\&NA; \; Fu \equiv PR\&NR; \; F1 \equiv PA\&NR;$$

## A. Data Collection

Previous section described the procedure of generating the data collection that the Central Unit of the Intelligent Iris Verifier Distributed System follows to interpret. It is a large-scale test in which the CU collects 52.759.182 intra-class and 10.991.943 inter-class scores from 1.441 simulated terminal stations. All these data are used for computing FAR and FRR curves (Fig. 1) that the CU follows to interpret according to a prescribed logical model in order to achieve the data understanding. FAR and FRR curves are reported for the first 100 (Test 1), 200 (Test 2) and 300 (Test 3) terminal stations, and finally for all 1.441 terminals (Test 4, Fig. 1).

On the intervals where FAR or FRR values are not known directly through experimental data, Pessimistic Odds of False Accept (POFA) and Pessimistic Odds of False Reject (POFR) are computed accordingly to the last known linear trends of FAR and FRR curves, respectively:

TABLE I. EXAMPLES OF FRR, FAR, AND POFA VALUES (FIG. 1)

| Threshold | FRR | FAR | POFA |
| --- | --- | --- | --- |
| 0,475 | 1,18E-6 | 3,79E-8 | - |
| 0,525 | 1,00E-4 | 0 | <1E-09 |
| 0,550 | 2,70E-4 | 0 | <1E-10 |

Now, let us comment around the knowledge K that CU follows to disseminate to all IIV terminals within its network and also on Fig. 3 from [1] who illustrates the way in which CU understands iris recognition and the space of binary iris codes. In Fig. 3 from [1] we see that the theory $T = (V, K)$ is a theory of creation for two planets, namely Imposter (IMP) and Genuine (GEN), allowed to form with material objects from VxV (with pairs of binary iris codes) and allowed to have satellites in such a way that those of IMP to stay out of any collision course with those of GEN. In this view, *evolutionary learning* (which is what an IIV do [1]) *means identifying 'planets' of knowledge / understanding (concepts) within the learned / classified data.*

Hence, evolving digital identities in an IIV is equivalent to dynamically controlling the system formed by IMP, GEN and their satellites in response to new enrollments (which expand the vocabulary V). The planet IMP / GEN is that core of the IMP / GEN zone in which the CU proves a crisp understanding of what it means to be *an imposter pair* / *a genuine pair*. If the IIV system is well trained, it knows an advanced stage of the process that forms IMP and GEN, hence the IIV sees their satellites as leftovers (of this formation process) that stay close to the core of their class, so close that the field of debris around IMP can't collide into the field of debris around GEN.

Obtaining too optimistic results is not our goal here, and therefore, in order to compensate for the fact that we use the same database on all terminals and in order to get a view over the worst-case scenario of our test, we allow the planet IMP to explode (to decrease in size and density with almost 100%) and to sent its material as satellites toward the confusion zone. Doing this is equivalent to a return in time along the history that formed IMP to an early stage where IMP was (perceived) just (as) a field of debris, a stage in which the knowledge of IIV about IMP was weaker than the knowledge illustrated in Fig. 3 from [1] (a stage in which IIV didn't proved a remarkable crisp understanding of what it means to be an imposter pair). With this occasion, we show another important difference between the *stationary* and the *evolutionary biometric systems*: the process of rolling back one direction of the system history is inconceivable in a stationary biometric system. Fig. 2 shows the numerical results collected from 1.441 terminals of an IIVDS, which obtained their local results by following this pessimistic scenario described above. What is truly remarkable is that even in this pessimistic scenario, IIVDS highlights the existence of a 'planet' of *absolute safety* within the space of genuine pairs: for 83,83% (9.214.655) of all genuine comparisons (10.991.943), IIVDS proves a crisp understanding of what it means to be a genuine pair (Fig. 2). Hence, a statistical decisional landscape [4] for classifying True Accepts could affect at most 16% (1.777.288) of the genuine pairs.

The next two sections shows the options that we had in endowing the IIV agent with a logic in which it understands its own experience during and after such complex tasks as the test described above.

III. INCONSISTENT DATA UNDERSTANDING

The firs attempt to find a logical formalization designed to be suitable for artificial understanding of the data illustrated in Fig. 1-2 started form the theoretical concept of Equal Error Rate (EER), which is the point (a threshold) where FAR and FRR equals each other. In our experience, we did not found such a crisp point. On the contrary, we found that the theoretical concept of EER point has a correspondence in a collection of EER points that may appear varying during different recognition tests, a collection that we called *a fuzzy EER interval*. In this view the fuzzy-logical decision landscape of iris recognition is given by three intervals: MPD (most probable different irides), EP (fuzzy equally probable or fuzzy EER interval), and MPI (most probable identical irides) such that:

$$\text{MPD} \cup \text{EP} \cup \text{MPI} = [0,1] \text{ and } \text{EP} = \text{MPD} \cap \text{MPI}.$$

If the logical state EP is observable during a functioning regime of the biometric system, there exist a binary iris code C and a digital identity I such that the positive claim P(C, I) or the negative claim N(C, I) commands the system to enter into this state. If the system is in EP state and accepts the positive claim P(C, I), and if the logic of recognition is 2-valent, then the propositional variable:

$$p = \text{``N(C, I) is false''}$$

*is true*. However, since the system is in EP state, also the propositional variable "N(C, I)" is true, and therefore:

$$p \equiv N(C, I),$$

and p reformulates as: $p \equiv$ "p is false", and it *is still true*. Hence, we managed to identify in the internal logic of the biometric system a propositional variable that is true and tells about itself that it is false. In other words, between the strings that are well-formed in the formal logic of the biometric system we found a non-empty support for the Liar Paradox. Therefore, the fuzzy 3-valent logical understanding of iris recognition described above is logically inconsistent [2] and will prove anything, sooner or later (this is the logical mechanism through which the *wolves* and the *lambs* [14] appear/enter in a stationary/non-adaptive biometric system, which in this way exceeds the framework of Consistent Biometry [1]). In conclusion, such a logical model is certainly not suitable for the IIV or for any other intelligent agent that aims to evolve *always to* and *always through* [1] logically consistent states.

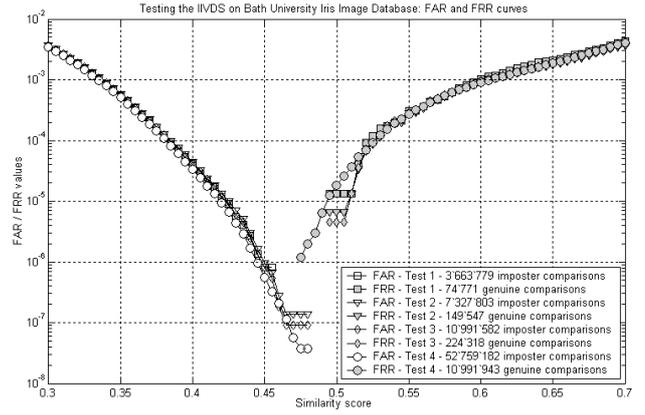

Figure 1. FAR and FRR curves in a large-scale test of IIVDS

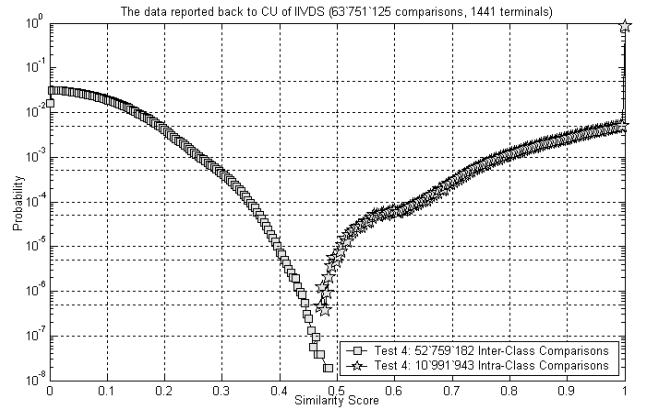

Figure 2. Inter-class (left) and intra-class (right) similarity scores obtained in a large-scale simulation with 1.441 terminal stations.

IV. ALMOST CONSISTENT DATA UNDERSTANDING

The second attempt to find a logical formalization designed to be suitable for artificial understanding of the data illustrated in Fig. 1-2 started form what we called *the dual concept of EER interval*, i.e. the *safety interval*. To illustrate this concept we give the following example: let us consider the intervals: I = [b, 1], D = [0, a], and O = (a, b), where b = $POFA^{-1}$(1E-10) ≈ 0.55, and a = $POFR^{-1}$(1E-10) ≈ 0.3725.

The interval O is a *safety interval* from the perspective of the biometric system, a *discomfort interval* from the perspective of the users, and an *uncertainty interval* from the perspective of logic. As specified in Section II, the meaning of the modal values of truth I, O, and D are:

D ≡ F0 ≡ PR&NA; O ≡ Fu ≡ PR&NR; I ≡ F1 ≡ PA&NR;

If D, O, and I are logical values of truth, they must reveal somehow their belonging to a Boolean algebra. Since we aim to find a consistent logic that models IIV understanding of the results illustrated in Fig. 1-2, the impossible state of the IIV biometric system will be encoded as PA&NA (which is exactly the modal state of truth EP from the previously attempted formalization), i.e. the state in which the biometric system accepts the positive and the nagative claims concerning the same iris code and the same digital identity (wolf-lamb pairs, [14]):

*Theorem 1 (N. Popescu-Bodorin, V. E. Balas):*

*If in an IIVDS the logic of accepts and rejects is the Propositional Binary Logic (PBL), then the state PA&NA is not observable for IIVDS (or in other words the IIVDS is logically controllable).*

The proof of the above theorem is almost done in the previous section where we showed that if such a state is observable, then the 2-valent logic of accepts and rejects is inconsistent, and this fact contradicts the hypothesis of our theorem because in PBL there is no room for the Liar Paradox.

The only problem now is to find a computational formalization of the Boolean algebra of intervals generated by E (the empty set), D, O, and I:

*Theorem 2 (N. Popescu-Bodorin, V. E. Balas):*

*The modal values of truth E, D, O, and I are four elements of a Boolean algebra defined over the congruence classes within $Z_8$ (modulo 8 integers). The intrinsic 8-valent logic of this Boolean algebra is the 8-vlaent formal logic language of computing with E, D, O, and I in a logically consistent manner.*

*Proof:* Let us consider the Boolean algebra generated by E, D, O, and I with the reunion, intersection and complement, $S = (\langle E, D, O, I \rangle, \cup, \cap, C)$, and let $\psi$ be the correspondence:

| $\psi:$ | E | D | O | I | OD | ID | IO | IOD |
|---|---|---|---|---|---|---|---|---|
| | 0 | 1 | 2 | 4 | 3 | 5 | 6 | 7 |

Let us consider that 'M' is an acronym for modulo and '$\geq$' is a legal logico-arithmetical operator that returns a binary value of truth. Let us consider the Boolean algebra $Z = (Z_8, p, s, n)$ where:

$\forall \hat{a}, \hat{b} \in \{\hat{0}, \hat{1}, \hat{2}, \hat{3}, \hat{4}, \hat{5}, \hat{6}, \hat{7}\}$:

$n(\hat{a}) = \hat{b} \Leftrightarrow b + a = 7,$

$p(\hat{a}, \hat{b}) = \hat{c} \Leftrightarrow c = \sum_{n=0}^{2} 2^n (aM2^{n+1} \geq 2^n)(bM2^{n+1} \geq 2^n),$

$s(\hat{a}, \hat{b}) = n(p(n(\hat{a}), n(\hat{b})))$.

The correspondence $\psi$ from above is an isomorphism between the Boolean algebras S and Z. A simple verification of this fact can be done by observing the similitude between the tables of the two sets of operations defined in S and Z, respectively. □

For IIVDS, the understanding of the results within Test 4 (Fig. 2) in the Boolean algebra Z is *almost consistent* because all construction here is based on the following assumption: *similarity scores 'a' and 'b' can't be reached through genuine and imposter comparisons, respectively* – and this affirmation is a statistical inference not a proved theorem. Hence the fact that '*PA&NA is not observable for IIVDS*' (Theorem 1) can be reformulated as '*PA&NA is almost certainly not observable for IIVDS*'.

## V. CONCLUSION

The large-scale test documented here (Test 4) showed that even in the pessimistic scenario described above in the Section II.A, IIVDS ensures *absolute safety* for 84% of accept cases. Despite the pressure artificially introduced in the space of imposter comparisons, the statistical aspect of recognition is so weak that the functioning regime specified by the thresholds 0.3725 and 0.55 (which define the safety interval) ensures for the IIVDS *outstanding performance* in terms of: **1E-10** pessimistic odds of false accept, **1E-10** pessimistic odds of false reject, **4.12E-4%** undecidable cases (2.7E-4% cases of honest positive claims and 1.42E-4% cases of honest negative claims), and a safety interval of width **0.1775** between the maximum reject and minimum accept scores. Hence, the IIVDS is an *almost consistent iris identifier*, at least.

ACKNOWLEDGMENT

We thankfully acknowledge the University of Bath and Prof. D. Monro for granting us access to the iris database.

REFERENCES

[1] N. Popescu-Bodorin, V.E. Balas, "Learning Iris Biometric Digital Identities for Secure Authentication", *Recent Advances in Intelligent Engineering Systems*, Springer Verlag, in press 2011.

[2] N. Popescu-Bodorin, V.E. Balas, "From Cognitive Binary Logic to Cognitive Intelligent Agents", Proc. 14th Int. Conf. on Intelligent Engineering Systems, pp. 337-340, IEEE Computer Society, May 2010.

[3] N. Popescu-Bodorin, V.E. Balas, "Comparing Haar-Hilbert and Log-Gabor based iris encoders on Bath Iris Image Database", Proc. 4th Int. Workshop on Soft Computing Applications (SOFA, 2010), pp. 191-196, IEEE Press, July 2010.

[4] J. Daugman, "Biometric Decision Landscapes", Technical Report No. TR482, University of Cambridge, 2000.

[5] J. Daugman, "New methods in iris recognition", IEEE Trans. Systems, Man, Cybernetics, B 37(5), pp 1167-1175, Oct. 2007.

[6] T. Tan, L. Ma, "Iris Recognition: Recent Progress and Remaining Challenges", Proc. of SPIE, Vol. 5404, pp. 183-194, Apr. 2004.

[7] L. Ma, T. Tan, Y. Wang, D. Zhang, "Personal Identification Based on Iris Texture Analysis", IEEE TPAMI, Vol. 25, No. 12, pp.1519-1533, 2003.

[8] S. Rakshit, D.M. Monro, "Robust Iris Feature Extraction and Matching", Proc. IEEE Int. Conf. on Digital Signal Processing, Jul. 2007.

[9] P. Grother, E. Tabassi, G. Quinn, W. Salamon, "Interagency report 7629: IREX I - Performance of iris recognition algorithms on standard images", N.I.S.T., Oct. 2009.

[10] D.M. Monro, S. Rakshit, D. Zhang, "DCT-Based Iris Recognition", IEEE TPAMI, Vol.29, No.4, pp. 586-595, 2000.

[11] Iris Challenge Evaluation, N.I.S.T., http://iris.nist.gov/ice/, Cited Mars 2011

[12] K.W. Bowyer, K. Hollingsworth, P.J. Flynn, "Image understanding for iris biometrics: a survey", Computer Vision and Image Understanding, vol. 110, no. 2, pp. 281-307, 2008.

[13] K.P. Hollingsworth, K.W. Bowyer, P.J. Flynn, The best bits in an iris code, IEEE TPAMI, Vol. 31, No. 6, pp. 964-973, June 2009.

[14] N. Yager, T. Dunstone, "The Biometric Menagerie", IEEE TPAMI, vol.32, no.2, pp.220-230, Feb. 2010.